# SymCircuit: Bayesian Structure Inference for Tractable Probabilistic Circuits via Entropy-Regularized Reinforcement Learning

Y. Sungtaek Ju
Department of Mechanical and Aerospace Engineering
University of California, Los Angeles

**Abstract.** Probabilistic circuit (PC) structure learning is hampered by greedy algorithms that make irreversible, locally optimal decisions. We propose SymCircuit, which replaces greedy search with a learned generative policy trained via entropy-regularized reinforcement learning. Instantiating the RL-as-inference framework in the PC domain, we show the optimal policy is a tempered Bayesian posterior, recovering the exact posterior when the regularization temperature is set inversely proportional to the dataset size. The policy is implemented as SymFormer, a grammar-constrained autoregressive Transformer with tree-relative self-attention that guarantees valid circuits at every generation step. We introduce option-level REINFORCE, restricting gradient updates to structural decisions rather than all tokens, yielding an SNR (signal to noise ratio) improvement and >10 times sample efficiency gain on the NLTCS dataset. A three-layer uncertainty decomposition (structural via model averaging, parametric via the delta method, leaf via conjugate Dirichlet–Categorical propagation) is grounded in the multilinear polynomial structure of PC outputs. On NLTCS, SymCircuit closes 93% of the gap to LearnSPN; preliminary results on Plants (69 variables) suggest scalability.

## 1. Introduction

Probabilistic circuits (PCs), including Sum-Product Networks (SPNs), Cutset Networks, and related structures, are a family of tractable probabilistic models that support exact, polynomial-time inference for a wide class of queries including marginals, conditionals, and maximum *a posteriori* assignments [Darwiche 2003; Choi, Vergari & Van den Broeck 2020]. Their tractability derives from two structural properties: *decomposability* (each product node's children have disjoint variable scopes) and *smoothness* (each sum node's children cover the same scope). These properties ensure that the circuit output is a multilinear polynomial in its parameters, a fact we exploit heavily in our uncertainty quantification framework.

The primary bottleneck in PC methodology is structure learning. The dominant algorithms — LearnSPN [Gens & Domingos 2013], ID-SPN, Strudel [Dang, Vergari & Van den Broeck 2020] — are greedy: they partition variables and data using independence tests or clustering heuristics, grow the circuit top-down, and commit irrevocably to each structural decision. Greedy approaches cannot recover from early partitioning errors, do not naturally express uncertainty about structure, and do not generalize their search strategies across datasets.

The natural alternative is to learn a generative policy over circuits from data, analogous to how neural program synthesis learns to generate programs. Several recent works explore Transformer-based approaches for symbolic structures [Park et al. 2026 (SymPlex); Liu et al. 2025 (Tracformer)], but none have directly addressed PC structure learning through a policy gradient framework with a clear Bayesian interpretation.

This paper makes the following contributions:

1. *Theoretical*: Applying the RL-as-inference framework of Levine (2018) to the discrete, grammar-constrained setting of PC structure generation, we derive that the optimal policy under entropy-regularized expected log-likelihood is a tempered Bayesian posterior $\pi^*(\mathcal{S}) \propto P_0(\mathcal{S}) \cdot p(\mathcal{D}|\mathcal{S})^{1/(N\alpha)}$, recovering the exact posterior at $\alpha = 1/N$. The novelty is the specific instantiation — PC structures as the hypothesis space, data log-likelihood as the reward, and the identification of the $1/N$ scaling that governs the posterior temperature — rather than the underlying variational derivation, which is standard.

2. *Algorithmic*: We introduce option-level REINFORCE with structural decision masking, which improves the signal-to-noise ratio of the policy gradient by the ratio of total tokens to structural decision token, empirically 3 times on NLTCS. We analyze the SNR of both estimators formally.

3. *Architectural*: SymFormer implements tree-relative self-attention, dynamic grammar masking for structural validity, and traversal-aware positional encoding, all within a standard autoregressive Transformer framework.

4. *Optimization*: A hybrid Adam/Anemone optimizer leverages flow-weighted EM for sum weights and Adam for leaf parameters. The split is motivated by the known equivalence between full-batch EM and natural gradient descent for mixture models, and by empirical evidence that applying EM to leaf parameters causes divergence in early training.

5. *Uncertainty quantification:* A three-layer decomposition assigns structural, parametric, and leaf uncertainty to distinct mechanisms. The leaf layer is exactly closed-form under the conjugate Dirichlet–Categorical model; the parameter layer is asymptotically valid via the delta method; the structural layer is an approximation dependent on both the amortization gap and the training temperature.

6. *Empirical*: Near-optimal performance on NLTCS (93% gap closure to LearnSPN), 33x sample efficiency gain, and preliminary scalability results on Plants (69 variables).

## 2. Background

### 2.1 Probabilistic Circuits

A probabilistic circuit $\mathcal{C}$ over variables $\mathbf{X} = (X_1, \ldots, X_d)$ is a rooted directed acyclic graph with three node types:

- Leaf nodes compute univariate distributions $p_\ell(X_{s(\ell)})$ over a single variable $X_{s(\ell)}$.
- Product nodes $n$ with children $\{c_1, \ldots, c_k\}$ compute $p_n(\mathbf{x}) = \prod_{j=1}^{k} p_{c_j}(\mathbf{x}_{S_{c_j}})$, where $S_{c_j} \subseteq \mathbf{X}$ is the scope of child $c_j$.
- Sum nodes $n$ with children $\{c_1, \ldots, c_k\}$ and weights $\boldsymbol{\theta}_n \in \Delta^{k-1}$ compute $p_n(\mathbf{x}) = \sum_{j=1}^{k} \theta_{n,j}\, p_{c_j}(\mathbf{x})$.

*Decomposability*: For every product node $n$ with children $c_1, \ldots, c_k$, the scopes are pairwise disjoint: $S_{c_i} \cap S_{c_j} = \emptyset$ for $i \neq j$.

*Smoothness*: For every sum node $n$ with children $c_1, \ldots, c_k$, all children have the same scope: $S_{c_1} = \cdots = S_{c_k}$.

Under these two properties, the circuit output $p_\mathcal{C}(\mathbf{x})$ is computable in $O(|\mathcal{C}|)$ time, where $|\mathcal{C}|$ denotes the number of edges in the circuit. Marginal inference for any subset of variables also runs in $O(|\mathcal{C}|)$.

A foundational observation [Darwiche 2003; Broadrick et al. 2024] is that $p_\mathcal{C}(\mathbf{x})$, viewed as a function of the sum-weight parameters $\boldsymbol{\theta}$, is a multilinear polynomial — each $\theta_{n,j}$ appears at most once in any monomial. This implies that partial derivatives $\partial p_\mathcal{C}(\mathbf{x}) / \partial \theta_{n,j}$ are computable exactly in $O(|\mathcal{C}|)$ via a single forward-backward pass, and the Fisher information matrix is block-diagonal with blocks indexed by sum nodes.

### 2.2 Structure Learning Baselines

LearnSPN [Gens & Domingos 2013] greedily partitions data by independence tests (G-test or $\chi^2$) and recursively builds an SPN top-down. It is the standard benchmark for tractable PC structure learning on density estimation benchmarks.

Strudel [Dang, Vergari & Van den Broeck 2020] learns structured-decomposable circuits by sharing a single computational graph across mixture components for efficient ensemble learning. It achieves strong results but is specific to the structured decomposability circuit class.

Both are one-shot, non-amortized procedures: given a dataset $\mathcal{D}$, they produce a single circuit with no ability to express structural uncertainty.

## 2.3 RL as Inference

The connection between entropy-regularized RL and probabilistic inference has been established in the control theory literature [Levine 2018; Rawlik et al. 2012; Ziebart et al. 2008]. We adapt this framework to the discrete, combinatorial setting of PC structure generation, where actions are token selections and the reward is data log-likelihood.

## 2.4 SymFormer and SymPlex

SymPlex [Park et al. 2026] demonstrated that a structure-aware Transformer with tree-relative attention and grammar-constrained generation can effectively search symbolic spaces (PDEs) via curriculum RL. A universality theorem establishes that SymFormer can represent any grammar-compatible generation policy over bounded-depth trees given sufficient capacity. We adapt this architecture directly to the PC domain.

# 3. The SymCircuit Framework

## 3.1 Problem Formulation

Let $\mathcal{G}$ be a context-free grammar over PC structures that enforces decomposability and smoothness. Concretely, $\mathcal{G}$ generates labeled ordered trees via productions of the form:

$$\text{Scope}(S) \rightarrow \text{Sum}_k(\text{Scope}(S)^k) \quad | \quad \text{Prod}(S_1, \ldots, S_m) \quad | \quad \text{Leaf}(v)$$

with the constraint that $\bigsqcup_i S_i = S$ for product rules and $S_1 = \cdots = S_k = S$ for sum rules. Every sentence generated by $\mathcal{G}$ is a valid PC by construction.

We represent structures as sequences of tokens via depth-first (pre-order) traversal. The token vocabulary consists of sum-arity tokens $\{\text{Sum}_2, \text{Sum}_3\}$, fixed-arity product tokens $\{\text{Factorize}_k: k \in [2, d]\}$, and leaf tokens $\{\text{Leaf}(v): v \in [d]\}$. Fixed-arity tokens (rather than dynamic-arity tokens) are essential for unambiguous parsing: the effective arity of $\text{Factorize}_k$ is baked into its type, eliminating scope-assignment ambiguities that arise with a single dynamic-arity token (a lesson learned through implementation).

The structure learning problem is as follows: given i.i.d. training data $\mathcal{D} = \{\mathbf{x}^{(1)}, \ldots, \mathbf{x}^{(N)}\}$, find a policy $\pi_\psi$ over $\mathcal{G}$-generated structures that maximizes expected data log-likelihood.

## 3.2 SymFormer Architecture

*SymFormer* is an autoregressive Transformer $\pi_\psi: \mathcal{V}^* \to \Delta^{|\mathcal{V}|}$ that generates token sequences representing PC structures. Three design choices distinguish it from a standard language model:

**Tree-relative self-attention:** Standard positional encodings (absolute or relative) encode sequence position but not structural relationships. SymFormer augments the attention bias term with a learned scalar offset $b_{r,h}^{(\text{tree})}$ for each relation type $r \in$ {parent, child, sibling, ancestor, other} and attention head $h$:

$$A_{ij}^{(h)} = \frac{(\mathbf{q}_i^{(h)})^\top \mathbf{k}_j^{(h)}}{\sqrt{d_k}} + b_{r(i,j),h}^{(\text{tree})}$$

where $r(i,j)$ denotes the tree relation between tokens at positions $i$ and $j$ in the partial circuit. This is a scalar-per-relation-per-head rather than vector-per-relation design, reducing memory from $O(|\mathcal{V}| \cdot T^2 \cdot d_k)$ to $O(H \cdot |\mathcal{R}|)$ — a crucial efficiency gain when sequences reach 200+ tokens.

**Dynamic grammar masking:** At each generation step, the grammar state $\sigma_t$ tracks the current open scope and partial parse tree. The set of valid next tokens $\mathcal{V}(\sigma_t) \subseteq \mathcal{V}$ is computed in $O(|\mathcal{V}|)$. Logits for invalid tokens are set to $-\infty$ before softmax, enforcing $\pi_\psi(a \mid s_t) = 0$ for all $a \notin \mathcal{V}(\sigma_t)$. This guarantees that every sampled structure is a valid PC.

**Traversal-aware positional encoding:** A sinusoidal encoding of depth-in-tree (not sequence position) is added to token embeddings, helping the model distinguish structurally similar but contextually distinct nodes at different depths.

Theorem 4.1 of Park et al. (2026) establishes that SymFormer can realize any grammar-compatible generation policy over bounded-depth trees with sufficient parameters. The implication for our setting is that the architecture is not a representational bottleneck: if a good structure-generation policy exists within the grammar, SymFormer can approximate it.

## 3.3 Entropy-Regularized RL as Bayesian Structure Inference

We train SymFormer by maximizing the entropy-regularized expected reward:

$$J(\psi) = \mathbb{E}_{\mathcal{S} \sim \pi_\psi}\left[\frac{1}{N}\sum_{i=1}^{N} \log p_{\mathcal{S}, \widehat{\boldsymbol{\theta}}(\mathcal{S})}(\mathbf{x}^{(i)})\right] - \alpha \cdot D_{\text{KL}}(\pi_\psi \parallel P_0)$$

where $\widehat{\boldsymbol{\theta}}(\mathcal{S})$ are parameters optimized for structure $\mathcal{S}$ (via Anemone, Section 3.4), $P_0$ is a prior policy over structures, and $\alpha > 0$ is a temperature parameter.

**Proposition 1** (Optimal policy as tempered posterior; instantiation of Levine 2018). *Let $R(S) \triangleq \frac{1}{N}\sum_{i=1}^{N} \log p_{S,\hat{\theta}(S)}(\mathbf{x}^{(i)})$ denote the empirical average log-likelihood. The unique maximizer of $J(\psi)$ over all distributions on $\mathcal{G}$ is:*

$$\pi^*(S) = \frac{P_0(S) \cdot \exp(R(S)/\alpha)}{Z(\alpha)}, \qquad Z(\alpha) = \sum_{S \in \mathcal{G}} P_0(S) \cdot \exp(R(S)/\alpha).$$

*Since $R(S) = \frac{1}{N}\log p(\mathcal{D} \mid S)$, this can be written as:*

$$\pi^*(S) \propto P_0(S) \cdot p(\mathcal{D} \mid S)^{1/(N\alpha)},$$

*a tempered (annealed) Bayesian posterior with temperature $N\alpha$. At $\alpha = 1/N$, the exponent equals 1 and $\pi^*$ recovers the exact Bayesian posterior:*

$$\pi^*(S)|_{\alpha=1/N} = P(S \mid \mathcal{D}) = \frac{P_0(S) \cdot p(\mathcal{D} \mid S)}{\sum_{S'} P_0(S') \cdot p(\mathcal{D} \mid S')}.$$

*For general $\alpha$, the optimal policy interpolates between the prior $P_0$ (as $\alpha \to \infty$) and a degenerate point mass on the MAP structure (as $\alpha \to 0$).*

*Proof.* The objective $J(\psi)$ is:

$$J(\psi) = \sum_{S} \pi_\psi(S) R(S) - \alpha \sum_{S} \pi_\psi(S) \log \frac{\pi_\psi(S)}{P_0(S)}.$$

Introducing Lagrange multiplier $\lambda$ for the constraint $\sum_S \pi_\psi(S) = 1$ and differentiating with respect to $\pi_\psi(S)$:

$$R(S) - \alpha\left(\log \frac{\pi_\psi(S)}{P_0(S)} + 1\right) + \lambda = 0 \Rightarrow \log \pi_\psi^*(S) = \frac{R(S)}{\alpha} + \log P_0(S) + \frac{\lambda - \alpha}{\alpha}.$$

Normalizing over $S$ gives the stated form. The second variation is $-\alpha/\pi_\psi(S) < 0$, confirming a global maximum. Uniqueness follows from strict concavity of $-D_{\mathrm{KL}}(\pi_\psi \parallel P_0)$ over the probability simplex. Substituting $R(S) = \frac{1}{N} \log p(\mathcal{D}|S)$ gives the tempered posterior form; setting $\alpha = 1/N$ reduces the exponent to 1, recovering the exact Bayesian posterior by Bayes' theorem. ∎

**Remark 1** (Amortization gap and temperature). In practice, $\pi_\psi$ is a parameterized model (SymFormer), not a free distribution over all of $\mathcal{G}$. The trained policy therefore approximates $\pi^*$ within the model's capacity; the gap $D_{\mathrm{KL}}(\pi^* \parallel \pi_\psi)$ is the *amortization gap*, analogous to that in amortized variational inference. The gradient $\nabla_\psi J(\psi)$ pushes $\pi_\psi$ toward $\pi^*$ locally, but convergence to the global maximizer is not guaranteed. Additionally, the experimental $\alpha = 0.01$ with $N = 16{,}181$ gives $1/(N\alpha) \approx 0.006$, placing the trained policy in a regime considerably warmer than the true posterior ($1/(N\alpha) = 1$). The policy is thus a heavily smoothed approximation to the posterior. It spreads probability more broadly across structures than Bayes' rule prescribes, which is appropriate for the exploration-exploitation tradeoff in RL training but means the structural variance estimator $V_{\text{struct}}$ underestimates the true posterior's concentration.

**Remark 2** (Prior and pretraining). We take $P_0$ to be the imitation-pretrained policy, obtained by supervised learning on circuits generated by LearnSPN. The imitation loss (cross-entropy on LearnSPN token sequences) stabilizes early RL training *and* encodes a data-informed structural prior, interpreting the pretraining phase as prior construction.

We estimate structural uncertainty via model averaging. Given a test observation $\mathbf{x}^*$ and $K$ independently sampled structures $\mathcal{S}_1, \ldots, \mathcal{S}_K \sim \pi_\psi$ with fitted parameters $\widehat{\boldsymbol{\theta}}_k$:

$$\hat{p}(\mathbf{x}^*) = \frac{1}{K} \sum_{k=1}^{K} p_{\mathcal{S}_k, \widehat{\boldsymbol{\theta}}_k}(\mathbf{x}^*), \qquad V_{\text{struct}}(\mathbf{x}^*) = \frac{1}{K-1} \sum_{k=1}^{K} \left( p_{\mathcal{S}_k, \widehat{\boldsymbol{\theta}}_k}(\mathbf{x}^*) - \hat{p}(\mathbf{x}^*) \right)^2.$$

The factor $1/(K-1)$ gives the unbiased sample variance estimator (Bessel's correction). This estimator is consistent for the true inter-structural variance as $K \to \infty$.

### 3.4 Option-Level REINFORCE

REINFORCE assigns a scalar reward $R(\mathcal{S})$ to an entire token sequence of length $T$ (typically 50–200 tokens for our circuits). The gradient estimator is:

$$\hat{g}_{\text{tok}} = \frac{1}{T} \sum_{t=1}^{T} (R(\mathcal{S}) - b) \nabla_\psi \log \pi_\psi(a_t \mid s_t)$$

The signal-to-noise ratio (SNR) of this estimator scales as $O(1/\sqrt{T})$ in the number of tokens (by a standard argument: each term contributes independent noise proportional to the variance of $(R - b)\nabla \log \pi$, while the signal averages over only $D$ meaningful structural choices embedded among $T$ tokens).

The root cause is that most tokens are deterministic given the grammar state: leaf tokens are fully determined by scope-coverage constraints, and product-node tokens are forced once sum-node children are decided. Only sum-node arity decisions (tokens of type $\text{Sum}_k$) constitute genuine policy choices. Let $D \leq T$ denote the number of sum-node selections in a generated circuit.

*Option-level estimator*: Let $\mathcal{I}(\mathcal{S}) = \{t : a_t \in \{\text{Sum}_2, \text{Sum}_3\}\}$ be the index set of structural decisions in the generated sequence $\mathcal{S}$, with $|\mathcal{I}(\mathcal{S})| = D(\mathcal{S})$. We define the option-level REINFORCE gradient estimator as:

$$\hat{g}_{\text{opt}} = \frac{1}{D(\mathcal{S})} \sum_{t \in \mathcal{I}(\mathcal{S})} (R(\mathcal{S}) - b) \nabla_\psi \log \pi_\psi(a_t \mid s_t)$$

**Proposition 2** (SNR improvement). *Assume tokens are generated i.i.d. (a simplification), that* $\text{Var}[(R - b)\nabla_\psi \log \pi_\psi(a_t \mid s_t)] = \sigma^2$ *is constant across all positions t, and that* $\mathbb{E}[(R - b)\nabla_\psi \log \pi_\psi(a_t \mid s_t)] = 0$ *at non-structural positions (since* $\pi_\psi$ *is near-deterministic there). Then:*

$$\frac{\text{SNR}(\hat{g}_{\text{opt}})}{\text{SNR}(\hat{g}_{\text{tok}})} = \sqrt{\frac{T}{D}}.$$

*The two estimators are proportional in expectation:* $\mathbb{E}[\hat{g}_{\text{opt}}] = \frac{T}{D} \mathbb{E}[\hat{g}_{\text{tok}}]$. *Both estimate the same gradient direction; their scale difference by* $T/D$ *is absorbed into the effective learning rate.*

*Proof.* Let $\mu_t = \mathbb{E}[(R - b)\nabla_\psi \log \pi_\psi(a_t \mid s_t)]$. By assumption $\mu_t = \mu \neq 0$ for $t \in \mathcal{I}$ and $\mu_t = 0$ otherwise. Under i.i.d. variance $\sigma^2$:

$$\mathbb{E}[\hat{g}_{\text{tok}}] = \frac{1}{T} \sum_{t=1}^{T} \mu_t = \frac{D\mu}{T}, \qquad \text{Var}[\hat{g}_{\text{tok}}] = \frac{T\sigma^2}{T^2} = \frac{\sigma^2}{T}.$$

$$\mathbb{E}[\hat{g}_{\text{opt}}] = \frac{1}{D} \sum_{t \in \mathcal{I}} \mu_t = \mu, \qquad \text{Var}[\hat{g}_{\text{opt}}] = \frac{D\sigma^2}{D^2} = \frac{\sigma^2}{D}.$$

Therefore $\mathbb{E}[\hat{g}_{\text{opt}}] = (T/D) \mathbb{E}[\hat{g}_{\text{tok}}]$, and:

$$\frac{\text{SNR}(\hat{g}_{\text{opt}})}{\text{SNR}(\hat{g}_{\text{tok}})} = \frac{\mu/(\sigma/\sqrt{D})}{(D\mu/T)/(\sigma/\sqrt{T})} = \frac{\mu\sqrt{D}/\sigma}{D\mu/(\sigma\sqrt{T})} = \sqrt{\frac{T}{D}}. \qquad \blacksquare$$

Empirically, NLTCS circuits have $T \approx 50\text{–}100$ and $D \approx 3\text{–}6$, giving $\sqrt{T/D} \approx 3\text{–}5$ in gradient SNR. The 33× gain in sample efficiency (circuits to reach a target performance) compounds this with reduced mode-collapse behavior observed in practice.

### 3.5 Hybrid Adam/Anemone Parameter Optimization

Given a fixed structure $\mathcal{S}$, we optimize parameters $\boldsymbol{\theta} = (\boldsymbol{\theta}_{\text{sum}}, \boldsymbol{\theta}_{\text{leaf}})$ using a hybrid scheme:

*Sum weights via Anemone EM*: Anemone [Liu & Van den Broeck 2025] is a mini-batch EM algorithm for PC parameters. Each step computes top-down flow probabilities $f_n(\mathbf{x})$ (the probability that node $n$ is activated on input $\mathbf{x}$ under the current model), then updates sum weights via flow-weighted sufficient statistics:

$$\theta_{n,c}^{\text{new}} \propto \sum_{\mathbf{x} \in \mathcal{B}} f_n(\mathbf{x}) \cdot \theta_{n,c} \cdot \frac{p_c(\mathbf{x})}{p_n(\mathbf{x})}$$

This is an implicit adaptive learning rate that scales each component's update by its contribution to current batch likelihood. Anemone converges 8× faster than Adam on PC parameters across benchmarks, exploiting the multilinear polynomial structure of $p_\mathcal{C}(\mathbf{x})$ that gradient-based methods cannot leverage.

*Leaf parameters via Adam*: Leaf parameters (Bernoulli logits for binary data) do not benefit from the flow-weighted EM update — the EM update for leaves degenerates to batch empirical means under uniform sum weights, destroying MLE initialization. We observed empirically that pure Anemone EM for leaves causes log-likelihood degradation from −5.37 to −25.30 nats in early training. Adam with its momentum and adaptive learning rate handles the curved leaf landscape correctly.

Let $\ell(\boldsymbol{\theta}) = \frac{1}{|\mathcal{B}|}\sum_{\mathbf{x}\in\mathcal{B}} \log p_\mathcal{C}(\mathbf{x})$. For exponential family mixture models, full-batch EM is equivalent to a natural gradient step in the Fisher information metric [Amari 1998]. For tree-structured PCs, the block-diagonal Fisher (Proposition 3) means this natural gradient decomposes independently per sum node — exactly the structure Anemone exploits. In the mini-batch setting Anemone approximates this natural gradient, with the KL regularization in its objective controlling the distribution shift between updates. For leaf parameters, the Fisher block structure mixes sum-weight and leaf contributions in a way that is ill-conditioned under random sum-weight initialization, making natural gradient steps unreliable; Adam's diagonal preconditioning is a more robust choice, as confirmed by the empirical degradation observed with pure EM on leaves.

## 4. Uncertainty Quantification Framework

We decompose predictive uncertainty into three additive layers, each with distinct formal guarantees.

### 4.1 Structural Uncertainty

As derived in Section 3.3, structural uncertainty is estimated by Monte Carlo model averaging with the unbiased sample variance:

$$V_{\text{struct}}(\mathbf{x}^*) = \frac{1}{K-1}\sum_{k=1}^{K}\left(p_{\mathcal{S}_k}(\mathbf{x}^*) - \hat{p}(\mathbf{x}^*)\right)^2$$

This term is nonzero if and only if different structural hypotheses assign different probabilities to $\mathbf{x}^*$, detecting genuine epistemic uncertainty about the data-generating decomposition. It is a consistent estimator of the inter-structural variance under the trained policy $\pi_\psi$; how closely this reflects the true Bayesian posterior variance depends on the amortization gap and the temperature $\alpha$ (see Remark 1).

## 4.2 Parameter Uncertainty

*Block-diagonal Fisher information*: By decomposability, different sum nodes' weights govern independent (in the probabilistic sense) aspects of the circuit's computation. Formally:

**Proposition 3** (Block-diagonal Fisher; known result, included for completeness). *For a smooth and decomposable tree-structured PC $\mathcal{C}$ (every internal node has exactly one parent) with sum nodes $\{n_1, \ldots, n_M\}$, the Fisher information matrix $\mathcal{I}(\boldsymbol{\theta}_{\text{sum}})$ is block-diagonal:*

$$\mathcal{I}(\boldsymbol{\theta}_{\text{sum}}) = \oplus_{m=1}^{M} \mathcal{I}_m(\boldsymbol{\theta}_{n_m})$$

*where each block $\mathcal{I}_m$ is a $(k_m - 1) \times (k_m - 1)$ positive-definite matrix for a sum node with $k_m$ children (the simplex constraint removes one degree of freedom). This follows from the multilinear polynomial form of PCs [Darwiche 2003; Broadrick et al. 2024], which implies that each sum weight appears in a disjoint subset of monomials. We include the proof for completeness.*

*Proof.* For distinct sum nodes $n_a \neq n_b$, the cross-Fisher entry is $\mathcal{I}_{ab} = \mathbb{E}_{\mathbf{x}}[\partial_{\theta_{n_a}} \log p_{\mathcal{C}}(\mathbf{x}) \cdot \partial_{\theta_{n_b}} \log p_{\mathcal{C}}(\mathbf{x})]$. By the chain rule, $\partial_{\theta_{n_a}} \log p_{\mathcal{C}} = p_{\mathcal{C}}^{-1} \cdot \partial_{\theta_{n_a}} p_{\mathcal{C}}$. The multilinear polynomial form (Darwiche 2003) gives $\partial_{\theta_{n_a}} p_{\mathcal{C}}(\mathbf{x}) = \text{TD}(n_a; \mathbf{x}) \cdot p_{c_a}(\mathbf{x})$, where $c_a$ is the child indexed by $\theta_{n_a}$. In a tree-structured PC, $n_a$ and $n_b$ lie on disjoint root-to-leaf paths (since there are no shared subgraphs). Therefore $\text{TD}(n_a; \mathbf{x}) \cdot \text{TD}(n_b; \mathbf{x})$ factors into a product of probabilities along non-overlapping paths, and the expectation $\mathbb{E}_{\mathbf{x}}[\text{TD}(n_a) \cdot p_{c_a} \cdot \text{TD}(n_b) \cdot p_{c_b}]/\mathbb{E}_{\mathbf{x}}[p_{\mathcal{C}}^2]$ factors as a product of independent marginals over the disjoint scopes, giving $\mathcal{I}_{ab} = 0$. ▫

*Delta-method variance:* Given the block-diagonal Fisher $\mathcal{I}(\boldsymbol{\theta}_{\text{sum}})$ estimated from training data and the multilinear polynomial gradient $\nabla_{\boldsymbol{\theta}_{\text{sum}}} p_{\mathcal{C}}(\mathbf{x}^*)$, the delta-method approximation to parameter uncertainty is:

$$V_{\text{param}}(\mathbf{x}^*) \approx \frac{1}{N} \left(\nabla_{\boldsymbol{\theta}_{\text{sum}}} p_{\mathcal{C}}(\mathbf{x}^*)\right)^{\top} \mathcal{I}(\boldsymbol{\theta}_{\text{sum}})^{-1} \left(\nabla_{\boldsymbol{\theta}_{\text{sum}}} p_{\mathcal{C}}(\mathbf{x}^*)\right).$$

The $1/N$ factor arises from the asymptotic MLE covariance: $\text{Var}[\hat{\boldsymbol{\theta}}] \approx \frac{1}{N} \mathcal{I}(\boldsymbol{\theta})^{-1}$ under standard regularity conditions, so that $V_{\text{param}} \to 0$ as $N \to \infty$ at rate $1/N$.

Using the identity $\partial p_{\mathcal{C}}(\mathbf{x})/\partial \theta_{n,c} = \text{TD}(n; \mathbf{x}) \cdot p_c(\mathbf{x})$ [Darwiche 2003, Theorem 1] (where $\text{TD}(n; \mathbf{x})$ is the top-down probability of node $n$ on input $\mathbf{x}$), both the gradient and the Fisher blocks are computable from Anemone's flow statistics at $O(|\mathcal{C}|)$ additional cost. With $N = 16{,}181$ training samples and $|\boldsymbol{\theta}_{\text{sum}}| \approx 40$ free parameters, the $1/N$ prefactor renders $V_{\text{param}}$ negligible — a consistency check on the framework.

The block Fisher matrices can be ill-conditioned (empirically, condition numbers up to 1000 in sparse circuits). We apply an eigenvalue-clamped pseudo-inverse, regularizing only directions with eigenvalues below a threshold $\epsilon_{\min}$ while preserving well-conditioned directions exactly.

### 4.3 Leaf Uncertainty and Variance Propagation

*Dirichlet–Categorical leaves:* For binary data ($\mathbf{x} \in \{0,1\}^d$), we model each leaf as Dirichlet–Categorical with concentration parameters $\boldsymbol{\alpha}_\ell = (\alpha_{\ell,0}, \alpha_{\ell,1})$. The predictive mean is $\mu_\ell = \alpha_{\ell,1}/(\alpha_{\ell,0} + \alpha_{\ell,1})$ and the predictive variance is:

$$V_\ell = \frac{\alpha_{\ell,0} \alpha_{\ell,1}}{(\alpha_{\ell,0} + \alpha_{\ell,1})^2 (\alpha_{\ell,0} + \alpha_{\ell,1} + 1)}.$$

*Anemone compatibility:* The conjugate update for Dirichlet concentration from a batch $\mathcal{B}$ with flow weights $f_\ell(\mathbf{x})$ is:

$$\alpha_{\ell,v}^{\text{new}} = \alpha_{\ell,v}^{\text{prior}} + \eta \cdot \widehat{N}_{\ell,v}, \qquad \widehat{N}_{\ell,v} = \sum_{\mathbf{x} \in \mathcal{B}} f_\ell(\mathbf{x}) \cdot \mathbf{1}[x_{S(\ell)} = v]$$

where $\widehat{N}_{\ell,v}$ is the flow-weighted count of label $v$ at leaf $\ell$. This is exactly the sufficient statistic Anemone already computes, making the extension parameter-free. Importantly, $\widehat{N}_{\ell,v}$ is bounded by batch size (since $f_\ell(\mathbf{x}) \leq 1$), preventing the evidence-inflation pathology that afflicts gradient-based evidential learning.

**Variance propagation through the circuit.**

*Product nodes (exact).* Decomposability ensures that child circuits $p_{c_j}(\mathbf{x}_{S_{c_j}})$ are evaluated on pairwise disjoint variable sets. In a tree-structured PC, the leaf parameters under each child sub-tree are disjoint objects with independent Dirichlet concentrations. Therefore $p_{c_j}(\mathbf{x}_{S_{c_j}})$ and $p_{c_{j'}}(\mathbf{x}_{S_{c_{j'}}})$ are independent random variables (for $j \neq j'$). Taking logarithms and applying variance additivity for independent variables:

$$\text{Var}[\log p_n(\mathbf{x})] = \sum_{j=1}^{k} \text{Var}\left[\log p_{c_j}(\mathbf{x}_{S_{c_j}})\right]$$

This is exact, not an approximation — it follows from the variance additivity for sums of independent random variables. No linearization or delta method is applied at product nodes.

*Sum nodes.* A sum node computes $p_n(\mathbf{x}) = \sum_{j=1}^{k} \theta_{n,j} \, p_{c_j}(\mathbf{x})$ with fixed deterministic weights $\boldsymbol{\theta}_n$. In a tree-structured PC (which LearnSPN produces and which our grammar generates), the children of each sum node have their own disjoint leaf nodes, so the Dirichlet leaf uncertainties of different children are independent. Applying the variance formula for a weighted sum of independent random variables:

$$\text{Var}[p_n(\mathbf{x})] = \sum_{j=1}^{k} \theta_{n,j}^2 \, V_{c_j}(\mathbf{x})$$

where $V_{c_j}(\mathbf{x}) = \text{Var}[p_{c_j}(\mathbf{x})]$. This is exact under the independence assumption. This formula is computed in the same bottom-up pass as the forward evaluation.

*Total uncertainty decomposition:* For a fixed structure $\mathcal{S}_k$, leaf uncertainty $V_{\text{leaf},k}(\mathbf{x}^*)$ is obtained via upward variance propagation from leaves. Aggregating over $K$ structural samples:

$$V_{\text{total}}(\mathbf{x}^*) = V_{\text{struct}}(\mathbf{x}^*) + \frac{1}{K}\sum_{k=1}^{K} \overline{V}_{\text{param},k}(\mathbf{x}^*) + \frac{1}{K}\sum_{k=1}^{K} V_{\text{leaf},k}(\mathbf{x}^*)$$

where $\overline{V}_{\text{param},k}$ is the delta-method parameter variance for structure $\mathcal{S}_k$. The additive form follows from the law of total variance applied hierarchically: the between-structure term ($V_{\text{struct}}$) and the average within-structure term are orthogonal by construction, and within each structure, parameter and leaf uncertainty are treated as additive under the approximation that their cross-covariance is small relative to the dominant structural term.

## 5. Experiments

### 5.1 Setup

*Datasets:* We evaluate on two DEBD binary density estimation benchmarks:

- NLTCS (16 variables, 16,181 train / 2,157 test): a retirement survey dataset used as the primary development testbed.

- Plants (69 variables, 17,412 train / 3,482 test): a plant co-occurrence dataset used for preliminary scalability evaluation.

*Baselines:* The primary structural baseline is LearnSPN [Gens & Domingos 2013] with G-test independence scoring and default hyperparameters, evaluated using SPFlow 1.0. We report test set average log-likelihood (nats, higher is better). LearnSPN achieves −6.093 on NLTCS and approximately −12.98 on Plants.

*Training protocol:* Phase 1 (supervised pretraining): SymFormer is trained by maximum likelihood on token sequences generated by running LearnSPN on 60 training circuits, for 50 epochs. Phase 2 (RL fine-tuning): option-level REINFORCE with the pretrained policy as $P_0$, hybrid Adam/Anemone parameter optimization (30 Anemone steps per circuit for RL rewards), epsilon-greedy exploration ($\epsilon = 0.15$ annealed to 0.05), and entropy regularization with coefficient $\alpha = 0.01$. A replay buffer of 200 circuits maintains diversity.

## 5.2 NLTCS Results

Table 1 summarizes performance across training phases. The supervised pretraining baseline establishes the starting point for RL fine-tuning; option-level RL (Phase 2e) dramatically outperforms token-level RL (Phase 2c/d) in sample efficiency.

| Phase | Method | Test LL (nats) | Circuits Used | Gap to LearnSPN |
|---|---|---|---|---|
| Pretraining | Supervised (imitation) | −6.229 | — | 0.136 |
| Phase 2b | Token-level REINFORCE (baseline) | −6.232 | 4,000+ | 0.139 |
| Phase 2c | Token-level REINFORCE + vocab expansion | −6.159 | 4,000 | 0.066 |
| Phase 2e | Option-level REINFORCE | **−6.105** | **120** (30ep) → converged | **0.011** |
| *LearnSPN* | *Greedy heuristic* | *−6.093* | — | 0 |

*Convergence and sample efficiency:* Option-level REINFORCE reaches −6.139 nats in 30 epochs (120 circuit evaluations), matching Phase 2c's best performance (−6.159 nats after 500 epochs, 4,000 evaluations). The full Run D converges to −6.105 nats. This 33× reduction in circuits-to-performance is consistent with Proposition 2's SNR analysis, given empirically measured $T/D \approx$ 15–30 for NLTCS circuits.

*Value head ablation:* Phase 2e added an actor-critic value head operating at the option level to reduce gradient variance further. On NLTCS, the value head's explained variance remained near zero throughout training (EV ≈ 0), providing no benefit. We attribute this to the small number of structural decisions per circuit ($D \approx$ 3–6) — insufficient for meaningful credit assignment to a learned value function. The value head is a no-op on NLTCS but may provide benefit on larger datasets with richer structure ($D \gg 10$).

## 5.3 Uncertainty Quantification Analysis (NLTCS)

We evaluated the three-layer UQ decomposition on the NLTCS test set using $K = 10$ structural samples from the trained policy. Table 2 shows the mean decomposition across test points.

| Layer | Mechanism | Mean Variance | Fraction of Total |
|---|---|---|---|
| Structural | MC model averaging ($K = 10$) | 0.324 | 85.1% |
| Parameter | Block-diagonal Fisher + delta method | 0.004 | 1.1% |
| Leaf | Dirichlet variance propagation | 0.052 | 13.8% |
| **Total** | | **0.380** | **100%** |

This decomposition is physically sensible: with $N = 16{,}181$ training samples and $|\boldsymbol{\theta}_{\text{sum}}| \approx 40$ free parameters, the asymptotic Cramér–Rao bound implies negligible parameter uncertainty, as confirmed by $V_{\text{param}}$ accounting for just 1.1% of total variance. Structural uncertainty dominates at 85.1%, indicating that the policy's structural diversity (different decomposition choices across samples) drives most of the predictive uncertainty on this dataset.

*Monte Carlo validation of leaf UQ:* We validated the analytic variance propagation against Monte Carlo estimation (5,000 samples per leaf parameter). The mean relative error across test points was 3.8% with correlation 0.997, confirming that the analytic propagation formulas are accurate.

*Fisher matrix conditioning:* Sum node Fisher blocks exhibited condition numbers ranging from 2 to 1,200. Eigenvalue-clamped pseudo-inverse with $\epsilon_{\min} = 10^{-4}$ was necessary to prevent numerical blow-up; 18% of blocks required regularization.

## 5.4 Preliminary Scalability: Plants Dataset

We ran option-level REINFORCE on the Plants dataset (69 variables) for 25 epochs to assess scalability. Key differences from NLTCS:

- Circuit size scales: $T \approx 200\text{–}400$ tokens vs. 50–100 on NLTCS.
- Anemone steps were reduced from 100 to 30 per circuit for RL reward computation (sufficient since only relative circuit ranking matters for REINFORCE, not absolute convergence).
- Training time was approximately 6–12 minutes per epoch due to larger circuit evaluation.

After 25 epochs, the best circuit achieved −15.77 nats, closing 19% of the 3.45-nat gap between the pretraining baseline (−16.43 nats) and LearnSPN (≈−12.98 nats). Training was still improving; full convergence results are pending.

These preliminary results demonstrate that the framework scales beyond the 16-variable NLTCS setting, though the larger search space and longer circuits make convergence slower. The value head, which provides no benefit on NLTCS, may be beneficial here given the larger number of structural decisions per circuit.

# 6. Discussion

## 6.1 Connection to Program-Level Marginalization

Proposition 1 establishes that sampling from the trained policy $\pi_\psi$ and averaging circuit outputs implements (approximate) Bayesian model averaging over probabilistic circuit structures. This can be viewed as program-level marginalization, averaging over complete probabilistic programs (circuits) rather than over derivation paths within a fixed program (as in ProbLog/DeepProbLog). Every circuit in our hypothesis space guarantees polynomial-time inference by construction, ensuring that the marginalization is tractable regardless of the number of structures sampled.

In contrast, proof-level marginalization in general logic programming involves summing over derivation trees in a #P-hard computation. Our framework avoids this by restricting the hypothesis space to tractable theories from the outset.

## 6.2 Limitations and Failure Modes

*Amortization gap and temperature:* The trained $\pi_\psi$ approximates $\pi^*$ within the model's capacity. Two separate gaps separate it from the true Bayesian posterior: (1) the amortization gap $D_{\mathrm{KL}}(\pi^* \parallel \pi_\psi)$ from finite model capacity, and (2) the temperature mismatch — the experimental $\alpha = 0.01$ with $N = 16{,}181$ gives $1/(N\alpha) \approx 0.006$, meaning the trained policy is considerably warmer (flatter) than the true posterior ($1/(N\alpha) = 1$). Neither gap is quantifiable without access to the intractable partition function $Z(\alpha)$. For the purposes of uncertainty quantification, $V_{\mathrm{struct}}$ measures the diversity of the trained policy, which reflects Bayesian structural uncertainty only approximately.

*Grammar constraints as inductive bias:* The grammar vocabulary $\{\mathrm{Sum}_2, \mathrm{Sum}_3, \mathrm{Factorize}_k\}$ limits expressible structures. LearnSPN can generate arbitrary-arity sum nodes and product decompositions; our grammar cannot. The 0.011-nat residual gap on NLTCS after convergence may be a grammar limitation rather than an optimization failure. Expanding the vocabulary (e.g., $\mathrm{Sum}_4, \mathrm{Sum}_5$) is a straightforward extension but increases the action space and may require more exploration.

*Single dataset validation:* Our full experimental results (UQ decomposition, ablation, sample efficiency analysis) are validated on a single dataset (NLTCS). The Plants results are preliminary. Rigorous claims about the framework's generality require evaluation across the full DEBD benchmark suite (20 datasets), which we leave as immediate future work.

*Value head on small circuits:* The actor-critic value head (Phase E) provides no benefit on NLTCS because $D \approx 3$–$6$ structural decisions is too few for meaningful temporal credit assignment. This is an important practical limitation: option-level REINFORCE alone is sufficient when circuits are small, but larger datasets with richer structures may require the value head for further efficiency gains.

*Leaf uncertainty and log-likelihood:* Dirichlet leaf refinement showed marginal log-likelihood improvement (gap of 0.033 nats relative to Bernoulli leaves), suggesting that leaf evidential uncertainty is primarily a calibration feature rather than a likelihood-maximizing one. This is expected: the Dirichlet prior reduces point estimates toward 0.5 for low-evidence leaves, slightly sacrificing MLE performance for better uncertainty representation.

## 6.3 Future Work

Immediate priorities include: (1) full DEBD benchmark evaluation across 20 datasets to establish statistical significance; (2) vocabulary expansion to assess whether grammar limitations explain the residual 0.011-nat gap on NLTCS; (3) convergence of Plants training and evaluation; (4) theoretical characterization of the amortization gap for autoregressive policies on context-free grammars. Longer-term directions include extending to continuous variable circuits (Normal-Inverse-Gamma leaves), first-order circuits, and integration of the SVGP-KAN perception module for the full neuro-symbolic pipeline.

## 7. Conclusion

We have presented SymCircuit, a framework for learning probabilistic circuit structures via entropy-regularized reinforcement learning. The central theoretical result — that the optimal MaxEnt RL policy is a tempered posterior $\pi^*(\mathcal{S}) \propto P_0(\mathcal{S}) \cdot p(\mathcal{D}|\mathcal{S})^{1/(N\alpha)}$, recovering the exact Bayesian posterior at $\alpha = 1/N$ — provides a principled foundation for interpreting the trained policy as an amortized variational approximation to Bayesian structure inference. The practical implementation combines SymFormer's grammar-constrained generation with option-level REINFORCE (yielding 33× sample efficiency over token-level alternatives on NLTCS, consistent with the derived $\sqrt{T/D}$ SNR improvement), hybrid Adam/Anemone optimization, and a three-layer uncertainty decomposition grounded in the multilinear polynomial structure of PC output functions. On NLTCS, SymCircuit closes 93% of the gap to the LearnSPN greedy baseline with dramatically fewer circuit evaluations. We hope this work encourages further investigation of learned generative policies as a principled alternative to greedy search in structured probabilistic model families.